# Probabilistic Temporal Reasoning with Endogenous Change

**Steve Hanks***
Dept. of CompSci and Engr
University of Washington
*hanks@cs.washington.edu*

**David Madigan**
UW Dept. of Statistics
and Fred Hutchinson CRC
*madigan@stat.washington.edu*

**Jonathan Gavrin**
UW Dept. of Anesthesiology
and Fred Hutchinson CRC
*jgavrin@u.washington.edu*

## Abstract

This paper presents a probabilistic model for reasoning about the state of a system as it changes over time, both due to exogenous and endogenous influences. Our target domain is a class of medical prediction problems that are neither so urgent as to preclude careful diagnosis nor progress so slowly as to allow arbitrary testing and treatment options. In these domains there is typically enough time to gather information about the patient's state and consider alternative diagnoses and treatments, but the temporal interaction between the timing of tests, treatments, and the course of the disease must also be considered.

Our approach is to elicit a qualitative structural model of the patient from a human expert—the model identifies important attributes, the way in which exogenous changes affect attribute values, and the way in which the patient's condition changes endogenously. We then elicit probabilistic information to capture the expert's uncertainty about the effects of tests and treatments and the nature and timing of endogenous state changes. This paper describes the model in the context of a problem in treating vehicle accident trauma, and suggests a method for solving the model based on the technique of sequential imputation.

A complementary goal of this work is to understand and synthesize a disparate collection of research efforts all using the name "probabilistic temporal reasoning." This paper analyzes related work and points out essential differences between our proposed model and other approaches in the literature.

## 1 Introduction

Our long-term goal is to support a human decision maker (physician) in diagnosing and treating a class of "temporal decision problems." Temporal decision problems are those in which the physician's options in gathering information about the patient or treating a diagnosed problem take roughly as much time as it takes the disease to cause significant changes in the patient's condition if left untreated. Therefore the physician will typically have some time to gather additional information, think about alternative diagnoses, or wait for a change in the patient's status. On the other hand, delaying treatment to gather information or to consider alternative diagnoses might lead to situations in which the patient's state changes for the worse.

We therefore need to model both the exogenous changes to the patient (traumatic events, treatments, tests) and endogenous changes (due to the progression of a disease or injury, for example), and we will concentrate on those problems in which endogenous and exogenous changes tend to occur at the same rate.[1]

The fact that our models will be elicited from a human expert requires that the model's structure and parameters be elicitable naturally and concisely. Our experience with elicitation leads us to a model in which the expert first provides a structural model of the domain, and then provides probabilistic parameters that quantify his uncertainty associated with that structure.

The paper is organized as follows: first we present a motivating example and describe the model. Next we discuss the inference problem, and describe techniques for incrementally building and solving the model. We finish with a comparison of this work to other research in the area, pointing out an essential difference between two disparate lines of work both using the name "probabilistic temporal reasoning."

## 2 Example

The following example will motivate the representational needs and choices for the model, and in Sec-tion 5 we will demonstrate how to solve a simplified version of this example.

---

*This work was supported in part by NSF grant IRI-9008670 and in part by a grant from the University of Washington Royalty Research Fund. Many thanks to Sandi Larsen for careful proofreading.

[1] *Exogenous* changes are those caused by forces external to the system as opposed to *endogenous* changes which are generated by forces internal to the system.

At time $t = 0$ an automobile accident occurs in which the patient, a healthy 45 year old man, is the driver. Contact with the steering wheel is noted. These collisions can be severe, moderate, or mild. Examination of the scene and prior experience suggests a moderate collision.

The usual *immediate* consequences of collisions of this sort are injuries to the head, abdominal cavity and internal organs, chest, and extremities. In this paper we will consider only head and chest injuries. Injury to the head can bruise the brain, which will cause it to begin swelling. Chest injuries can include a fractured sternum, one or both punctured lungs, and bleeding in the chest cavity.

The collision itself can be modeled as an exogenous event, which can instantaneously cause certain changes in the patient's state: trauma to the brain, broken sternum, punctured lung, and bleeding in the chest cavity.

These (instantaneous) state changes initiate a set of "processes" that will generate subsequent state changes. For example, brain trauma will cause the brain to begin swelling, which tends to increase intracranial pressure, which in turn will eventually cause dilated pupils and loss of consciousness. Internal bleeding decreases blood volume over time, which tends to destabilize vital signs (pulse and blood pressure). Bleeding into the chest cavity will also eventually increase pressure on the heart, decreasing its efficiency, further destabilizing vital signs. Impaired lung function decreases oxygen transfer to the tissues, which will eventually compromise the heart's functioning, and lead to a deficient oxygen supply to the brain, eventually manifesting itself as light-headedness and loss of consciousness.

These changes are not immediate, and to reason about the situation we must provide additional information about how quickly and under what circumstances the changes will occur. In many cases several processes (e.g. decreasing blood volume, increasing pressure on the heart) jointly influence a state variable (in this case vital signs).

Suppose that at time $t = 10$ minutes the patient is observed by paramedics. They observe a probable broken sternum, the patient is complaining of shortness of breath and dizziness, vital signs are unstable, but pupils are not dilated. From this we should be able to reason backward that if the collision had caused a head injury, the pupils would be dilated. At the same time, the probable chest injury and unstable vital signs suggest internal bleeding, which will soon cause serious problems if left unattended. Intravenous fluids should probably be administered immediately to increase blood volume, and if transportation to the hospital is expected to take more than 20 minutes, it might be best to insert a chest tube to drain blood from the chest cavity and reduce pressure on the heart. Finally, the collision and shortness of breath suggest a collapsed lung and decreased oxygen transfer, which should be treated immediately by administering oxygen.

The rest of this paper will develop a model that will allow us to reason about such a dynamic system where changes are due both to exogenous events (collisions, treatments, observations) and endogenous effects (internal bleeding indirectly causing unstable vital signs). We begin by developing a formal characterization based on semi-Markov models, then introduce structural and simplifying assumptions that aid in the elicitation and solution process. We then suggest a simulation approach to solving the model, which we demonstrate on a simplified version of the scenario presented above.

## 3 Attributes and Events

We will describe the system's[2] state in terms of a vector of *attributes*, each of which has a set of possible *values*. The system's state at a single point in time is fully specified by the assignment of a value to each attribute. We will use $A_i$ to refer to the $i^{th}$ attribute $i = 1, 2, \ldots, m$, and $v_{i,j}$ $j = 1, 2, \ldots, m_i$ to refer to the set of values it can take on.

In our simple example the attributes, their abbreviations, and their values are:

1. Severity of the collision (SC) (mild, moderate, strong)
2. Internal bleeding in the chest cavity (IB) (none, slight, gross)
3. Head injury (HI) (false, true)
4. Pupils dilated (PD) (false, true)
5. Vital signs (VS) (normal, unstable, flat)

Since we need to keep track of the system's state as it changes over time, a system state is to be interpreted with respect to a time point $t$. In this paper we will assume a discrete model of time, using $\delta$ to refer to the smallest possible time increment.

We will also need to keep track of the direction of change of each attribute. Each attribute/value pair will be annotated to indicate whether it is increasing, steady, or decreasing. We will use the symbols ↑ ↓ and ↔ for these three values, so (IB, slight, ↑) means that internal bleeding is currently slight, but without intervention will eventually become gross. Note that direction of change implies an *ordering* over values for each attribute. Without loss of generality we will assume that $v_{i,1} \prec v_{i,2} \prec \ldots v_{i,n}$.

---
[2]We will refer in the abstract to the "system" and in the context of the example to the "patient." We mean the two terms interchangeably.

## 3.1 Exogenous events

An *event* generally refers to an instantaneous change in (or observation of) the system's state. An exogenous event is an event that is due to some force applied from outside the system. The collision itself, any treatment like administering oxygen or ingesting a drug, and even tests and observations are exogenous events.

Events are probabilistic transformations from a system state at one point in time $t$, the time at which the event occurs, to the "next instant" $t + \delta$, the time at which the event's effects are realized.

The effects of an event depend on the values of relevant attributes at time $t$ but not the direction of change, so we define an event as a probabilistic mapping from vectors of pairs $(A_i, v_{i,j})$ to a probability distribution over vectors describing the system's state at $(t + \delta)$.

Our representation takes an event to be a set of *consequences* of the form

$$\{(d_1, p_{1,1}, c_{1,1}), (d_1, p_{1,2}, c_{1,2}), \ldots, (d_1, p_{1,n_1}, c_{1,n_1}),$$
$$(d_2, p_{2,1}, c_{2,1}), (d_2, p_{2,2}, c_{2,2}), \ldots, (d_2, p_{2,n_2}, c_{2,n_2}),$$
$$\vdots$$
$$\ldots \qquad (d_k, p_{k,n_k}, c_{k,n_k}) \}$$

where $d_i$ is a *conditioning* or *state description* expression (a boolean combination of attribute/value pairs), the $p_{i,j}$ values are probabilities, and the $c_{i,j}$ describe *changes* in attribute values effected by the event. Each $d_i$ is (certainly) either true or false with respect to any state $s$, and we require that for any event description the $d_i$ be mutually exclusive and exhaustive and that $\sum_j p_{i,j} = 1$ for all $i$. In this way we ensure that an event describes a probability distribution over $c_{i,j}$ sets.

Each $c_{i,j}$ is a set of attribute/value pairs indicating the *new* value for that attribute that is (instantaneously) realized. That is, $c$ is a set of pairs $(a_i = v_i)$ except that it need not dictate a value for all the attributes in the state space. In the case that an attribute $a_j$ is absent from $c$, the attribute is assumed to retain its previous value.

This representation is formally equivalent to a Markov transition matrix, but is potentially more compact because the event needs to mention in $d_i$ only those attributes that are relevant to its effects and in the $c_{i,j}$ only those attributes that it changes. It is described in more detail in [Hanks, 1990] and [Kushmerick et al., 1995].

In our simple example there are only three events: the initial collision at $t = 0$ and the two subsequent observations of the patient at $t = 10$. Suppose that we have gathered the following statistics about these accidents:

- If the accident is mild, then the probability that head injury occurs is 0.01 and the probability of an injury resulting in internal bleeding (none, slight, gross) is $(0.8, 0.15, 0.05)$.

- If the accident is moderate, then the probability that head injury occurs is 0.1 and the probability of an injury resulting in internal bleeding (none, slight, gross) is $(0.5, 0.4, 0.1)$.

- If the accident is severe, then the probability that head injury occurs is 0.25 and the probability of an injury resulting in internal bleeding (none, slight, gross) is $(0.3, 0.5, 0.2)$.

The collision event can then be described by a set of 18 consequences (one for each possible value of the three relevant attributes):

((ACC=MILD), .008, (HI=TRUE, IB=NONE))
((ACC=MILD), .0015, (HI=TRUE, IB=SLIGHT))
((ACC=MILD), .0005, (HI=TRUE, IB=GROSS))
((ACC=MILD), .792, (HI=FALSE, IB=NONE))
((ACC=MILD), .1485, (HI=FALSE, IB=SLIGHT))
((ACC=MILD), .0495, (HI=FALSE, IB=GROSS))
((ACC=MODERATE), .05, (HI=TRUE, IB=NONE))
((ACC=MODERATE), .04, (HI=TRUE, IB=SLIGHT))
((ACC=MODERATE), .01, (HI=TRUE, IB=GROSS))
((ACC=MODERATE), .72, (HI=FALSE, IB=NONE))
((ACC=MODERATE), .135, (HI=FALSE, IB=SLIGHT))
((ACC=MODERATE), .045, (HI=FALSE, IB=GROSS))
((ACC=SEVERE), .075, (HI=TRUE, IB=NONE))
((ACC=SEVERE), .125, (HI=TRUE, IB=SLIGHT))
((ACC=SEVERE), .05, (HI=TRUE, IB=GROSS))
((ACC=SEVERE), .225, (HI=FALSE, IB=NONE))
((ACC=SEVERE), .375, (HI=FALSE, IB=SLIGHT))
((ACC=SEVERE), .15, (HI=FALSE, IB=GROSS))

This representation easily can be extended to handle observations: we add an *observation* variable $o_{i,j}$ to each element of the event's consequence set,

$$E = \{\ldots (d_i, p_{i,j}, c_{i,j}, o_{i,j}) \ldots\},$$

where one of the $o_{i,j}$ will be reported when $E$ occurs. The observation $o_{i,j}$ will be reported if and only if the occurrence of $E$ results in the realization of its $(i, j)^{th}$ consequence. Information about which consequence was realized provides information about the system's state by providing information both about $d_i$ (the pre-execution state) and about $c_{i,j}$ (the changes made to that state).

The information an event supplies when it occurs can be ambiguous, however, since first some attributes might not appear in the $d_i$ or $c_{i,j}$ sets, thus the event will provide no information about that attribute,[3] and second, the same observation can be assigned to more than one consequence, in which case the agent is uncertain as to which of those consequences occurred. In the extreme, an event that has the *same* observation attached to *all* its consequences provides no information about the world. That is the case with the collision event: it changes the world but provides no direct information about what changes it effected.

---

[3] Actually an event can provide information about the value of an unmentioned attribute $a$ indirectly if it provides direct information about the value of some other attribute $a'$, and if $a$ and $a'$ are correlated.

In the example there are two possible observations we can make, which provide information about vital signs and pupil dilation respectively. Suppose that vital-sign information can always be gathered accurately, but that judgments about pupil dilation can be mistaken; in particular the paramedic reports dilated if the pupils appear dilated and ok if they do not, but P(dilated|PD = false) = .25 and P(ok|PD = true) = .10. These two events (neither of which change the system's state at all) are represented as follows:

{ ((VS=NORMAL), 1.0, (), NORMAL),
  ((VS=UNSTABLE), 1.0, (), UNSTABLE),
  ((VS=FLAT), 1.0, (), FLAT)) }

{ ((PD=TRUE), 0.9, (), DILATED),
  ((PD=TRUE), 0.1, (), OK),
  ((PD=NO), 0.75, (), OK),
  ((PD=NO), 0.25, (), DILATED) }

This model of informational actions is equivalent in expressive power to the definition of actions and information in Partially Observable Markov Decision Processes [Monahan, 1982] and is described in more detail in [Draper et al., 1994].

This event model can be used to build graphical structures that can in turn be used to reason about temporal problems. In our example we have three events: collision C, observe vital signs OVS and observe pupil dilation OPD. Suppose the first happened at time $t = 0$, and the second and third happened in quick succession starting at $t = 10$. The variable CS refers to "collision severity," which can take on values mild, moderate, or severe as described above. By rights it should be part of the temporal state, but we omit it for the sake of brevity because it does not figure in analysis of the system after $t = \delta$.

Figure 1 depicts the model so far, which can be built entirely from the event descriptions above along with priors on the initial values of the state variables. Priors for CS were given above, and we will assume that HI is initially false, IB is initially false, VS is initially normal and PD is initially false. The domain expert could provide different priors if these turned out to be unrealistic.

The model so far does not describe the behavior of the system between $t = \delta$ and $t = 10$, however. For that we need to develop a model of *endogenous change* that will allow us to predict changes in the system's state that occur between the times of known exogenous changes.

## 4 Endogenous change

Our model of endogenous change is built around expert-elicited rules describing situations like:

- If a head injury occurs, the brain will start to swell, and if left unchecked the swelling will cause the pupils to dilate within 3 to 7 minutes.

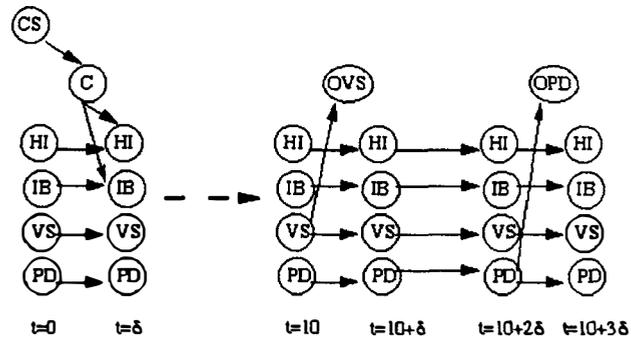

Figure 1: A Discrete-Event Model

Figure 2: Individual endogenous influences

- If internal bleeding begins, the blood volume will start to fall, which will tend to destabilize vital signs. The time required to destabilize vital signs will depend on the severity of bleeding: if the bleeding is slight, it will take between 30 and 60 minutes; if the bleeding is gross, it will take from 2 to 5 minutes.

- A head injury also tends to destabilize vital signs, taking between 2 and 5 minutes to make them unstable, and 10 minutes to 10 hours to make them flat.

These rules can be looked on as defining tables relating values of an "influencing" variable to transitions in an "influenced" variable. Figure 2 shows tables for the three rules informally described above. This example uses intervals interpreted as uniform probability distributions for the information about transition times, but other information (e.g. normal distributions) could be used instead. Note that these rules assert (with certainty) that the change *will occur* if the system is not changed in the meantime. To capture the case in which an endogenous change might or might not occur, the right interval endpoint can be made arbitrarily large.

One problem we still have to confront is the fact that the rules represent "local" influences on an attribute: implicitly the expert is saying that slight internal bleeding will destabilize vital signs in 30 to 60, minutes *all other things being equal*, but there may be other forces acting on the attribute simultaneously. In general, influences will be either *concordant*, all pushing the variable either higher or lower, or *contrary*, pushing the variable in the opposite direction. We need to be able to combine a set of influences into a single direction of change and estimated time to transition.

We do so in two stages, first combining concordant positive and negative influences to produce a single estimated transition time to the next higher and lower values, then combining contrary influences to produce a single direction and estimated transition time. For both combinations we have adopted very simple and arbitrary linear models, which nonetheless seem to agree closely with our domain expert's intuition on sample scenarios.

First consider combining two positive influences[4] on the same attribute, where the first predicts a change in $a$ time units, the second predicts a change in $b$ time units, and $a < b$. The actual time to transition should therefore be less than $a$, decreasing as $b \to a$ and approaching $a$ as $b \to \infty$. We stipulate a linear model with the following additional constraints:

- if $a = b$ then the transition time is $\frac{a}{2}$,
- if $b > 100a$ then the transition time is $a$

which leads us to a combination function of the form

$$f(a,b) = \begin{cases} f(b,a) & \text{if } b < a \\ a & \text{if } b > 100a \\ \frac{a}{2} + \frac{(b-a)}{198} & \text{otherwise.} \end{cases}$$

To combine contrary influences consider the case in which the first is a negative influence with time estimate $a$ and the second is a positive influence with time estimate $b$, and $a < b$. In this case the transition should be to the next negative value, but it should take *longer* than $a$ (due to $b$'s offsetting influence). In the limit, as $b \to a$ the transition should take arbitrarily long (the two influences offset), and as $b \to \infty$ the transition time should be close to $a$. Once again we stipulate a linear model with some additional constraints:

- if $a = b$ the transition time is $100a$
- if $b \geq 100a$ the transition time is $a$

which leads us to a combination function of the form

$$g(a,b) = \begin{cases} g(b,a) & \text{if } b < a \\ a & \text{if } b > 100a \\ 100a - (b-a) & \text{otherwise.} \end{cases}$$

In both cases the expert can deliver imprecise estimates for transition times—we are using intervals of the form $(a_1, a_2)$ and $(b_1, b_2)$ to capture that imprecision. We can apply both functions in a straightforward manner to compute aggregate influences given interval transition times. In the case of concordant influences for example, the shortest possible transition time consistent with expert judgment comes when $a = a_1$ and

---

[4]The same analysis holds for negative influences.

| HI/IB= | N/N | N/S | N/G | Y/N | Y/S | Y/G |
|---|---|---|---|---|---|---|
| VS= N | ↔ | ↓ [30,60] | ↓ [2,5] | ↓ [2,5] | ↓ [1.93,4.91] | ↓ [1,2.5] |
| U | ↔ | ↓ [60,120] | ↓ [10,20] | ↓ [10,600] | ↓ [9.8,119.75] | ↓ [5,19.97] |
| P | ↔ | ↔ | ↔ | ↔ | ↔ | ↔ |

Figure 3: Aggregated endogenous influences

$b = b_1$ and the longest consistent transition time occurs when $a = a_2$ and $b = b_2$. Therefore

$$f((a_1, a_2), (b_1, b_2)) = (f(a_1, b_1), f(a_2, b_2)).$$

Returning to the example we see a single concordant influence—both HI and IB have a potential negative influence on VS—and no contrary influences. We can therefore compute a table that captures their joint influence, which appears in Figure 3.

We should stress that this is a very simple and arbitrary model of endogenous change. First of all it mixes the idea of "force" with the idea of "transition time." A more physically realistic model might talk of the "destabilizing force" that head injury and blood loss have on vital signs, then talk in a more principled way about how these forces interact and how they affect time to transition. Our model in effect assumes that these "forces" are being elicited and compared using the same set of units for all relevant attributes. Second, our choice of a linear model was made only for the sake of simplicity—for any system there are obviously more sophisticated and realistic methods for modeling the effects of combined influences. We are encouraged, however, by the predictive power of our model even with these extreme assumptions: the domain expert is generally satisfied with the transition times that result from aggregating with these rules, and has yet to see a case in which computations performed on the model depend significantly on the exact aggregation method used. In any event we can always produce arbitrarily complex models of concordant and contrary influences by directly assessing aggregated influence models like the one in Figure 3.

We have now completed the model begun in Figure 1 by defining transition probabilities for an interval of time between the occurrence of exogenous events. Now we store three pieces of information for each attribute: its value, its direction of change, and its estimated time to transition. To reason about the distribution over the interval from $t_1 + \delta$ to $t_2$ (where $t_1$ and $t_2$ are both times at which exogenous events occur) we begin with the distribution over states generated by the $t_1$ exogenous event, then update transition direction and transition time for those attributes that were *changed* by that event. Then there will be a set of (probability distribution over) next possible endogenous state changes. Assuming every endogenous change takes non-zero time, there eventually will be a time at which there are *no* such changes (in which case we will have computed the distribution over states at $t_2$). Otherwise we can treat every possible next event exactly as an *exogenous* event: we update the state according to the attribute value it changes, update the expected transition times for the attributes that do not change value, and repeat.

It is difficult to capture this inference process in a fixed graphical network like Figure 1 since the number of endogenous events that occur between successive exogenous events is not fixed ahead of time. For that reason we have developed a simulation method for solving the model that handles both exogenous and endogenous updates. We now turn to the inference problem and a computational solution.

## 5 The Inference Problem

We now confront the prediction and diagnosis problems: given a set of exogenous events along with the times at which they occur, we want to make predictions about future states of the system as well as reason about the past and present states of unobservable aspects of the system. In our example we might want to know (1) the current extent of internal bleeding, since severe bleeding might demand immediate treatment, (2) the severity of the original collision, and (3) the effects of delaying treatment for some period of time.

We already noted a number of ways to compute these probabilities. First we could explicitly generate all temporal trajectories and associated probabilities, from which we could then recover the joint distribution over system states at each point in time. There are obvious space and time problems with the approach in that the number of trajectories will tend to grow exponentially.

A graphical model like the one pictured in Figure 1 offers a more compact representation, but the problem of how to supply transition probabilities for the endogenous events (i.e. transition probabilities between successive exogenous events) remains. One solution would be to perform a closed-form analysis of the possible changes in the interval, coming up with probabilities of the form $P(s_j \text{ at } t_{i+1}|s_i \text{ at } t_i)$ for all states $s_i$ and $s_j$ and all pairs of successive exogenous events occurring at $t_i$ and $t_{i+1}$. The problem with this approach is that for each interval it requires reasoning about transitions from *every possible* input state, even though many of those will not in fact occur.

We are currently exploring stochastic simulation methods for solving the system, since they tend to be reasonably space efficient, focus attention on relatively likely scenarios, and support real-time problem solving in the sense that they can be interrupted and provide sketchy information if time constraints demand it. Our current implementation employs the technique of *sequential imputation* [Kong et al., 1994], which has the additional advantage that we can extend the scenario (by adding new exogenous events) without having to discard the results of previous trials.

### 5.1 Sequential Imputation

Sequential imputation is an importance sampling technique that allows for incremental absorption of new observations. For expository purposes we sketch the method for two consecutive observations, $y_1$ and $y_2$, and corresponding attributes $a_1$ and $a_2$. Let $\Delta$ denote the quantity we want to reason about. $\Delta$ could be $a_2$ for instance, or some future $a_3$, or, we might want to reason backward in time and make inferences about $a_1$. In fact $\Delta$ could be an arbitrary function of any of the variables. To perform this inference, we need to compute $\Pr(\Delta \mid y_1, y_2)$. In general this will be intractable. However, we can re-express $\Pr(\Delta \mid y_1, y_2)$ as an expectation and then evaluate the expectation arbitrarily accurately using importance sampling:

$$\Pr(\Delta \mid y_1, y_2) = \int \Pr(\Delta \mid y_1, y_2, a_1, a_2) \Pr(a_1, a_2 \mid y_1, y_2) da_1 da_2.$$

Simple Monte Carlo evaluation of the above integral requires that we simulate from $\Pr(a_1, a_2 \mid y_1, y_2)$, which will typically be intractable. Instead, we draw $a_1^i$ from $\Pr(a_1 \mid y_1)$ and then, $a_2^i$ from $\Pr(a_2 \mid y_1, a_1^i, y_2)$ and maintain the veracity of the approximation with importance sampling weights. It may be necessary to appeal to stochastic algorithms to draw from these distributions, but the key point is that each involves simulating forward in time from complete information.

Now, we estimate $\Pr(\Delta \mid y_1, y_2)$ as:

$$\Pr(\Delta \mid y_1, y_2) \approx \sum_{i=1}^{n} w_i \Pr(\Delta \mid y_1, y_2, a_1^i, a_2^i)$$

where the importance sampling weights are:

$$w_i = \frac{\Pr(a_1^i, a_2^i \mid y_1, y_2)}{\Pr(a_1^i \mid y_1)\Pr(a_2^i \mid y_1, a_1^i, y_2)} = \frac{\Pr(y_2 \mid y_1, a_1^i)}{\Pr(y_2 \mid y_1)}.$$

Crucially, since $\Pr(y_2 \mid y_1)$ is the same across the imputations, it is absorbed in the normalization of the weights.

If we add a third observation, $y_3$, we draw $a_3^i$ from $\Pr(a_3 \mid y_3, a_2^i, y_2)$, and the weights become:

$$w_i = \frac{\Pr(y_2 \mid y_1, a_1^i)\Pr(y_3 \mid y_2, a_2^i, y_1, a_1^i)}{constant}.$$

Note that although this does *not* require that we generate new realizations from the first two time points, it does provide an estimate of, for instance $\Pr(a_1 \mid y_1, y_2, y_3)$.

### 5.2 Application to the temporal model

Now we apply the technique to our temporal model, assuming a sequence of events $E_1, E_2, \ldots, E_n$ occurring at known times $t_1, t_2, \ldots, t_n$ and a set of (actual) observations $o_1, o_2, \ldots, o_m$. We also need to specify a set $Q$ of query propositions of the form $(A_i, t_j)$, indicating that we are ultimately interested in attribute $A_i$'s value at time $t_j$.

In addition we must determine the set of attributes/time pairs required to compute the importance sampling weight, but this can be done directly by examining the events that generate the observations: if observation $o_i$ was generated by event $E_j$ then we need to know the value of every $(A_k, t_j)$ where $A_k$ is an attribute that appears in any consequence of $E_j$ that mentions $o_i$. Let **W** be this set of attributes relevant to the importance weight.

Trial generation then proceeds as follows:

1. An initial state is generated using the prior distribution provided by the user. Change directions for all attributes are set to $\leftrightarrow$. The current time $t_c$ is initialized to $t_1$.
2. Repeat for all exogenous events $E_1, E_2, \ldots, E_n$:
   (a) If $i$ is the current event and $(A_j, t_i) \in \mathbf{Q} \cup \mathbf{W}$ then record the value of $(A_j, t_i)$
   (b) To simulate the exogenous event $E_i$, we find the (unique) consequence of $E_i$ true in the current state; we sample from the distribution associated with that consequence, and apply the sampled change set to the current state.
   (c) Re-compute the new influences exerted on the state by the attributes that just changed state.
   (d) Subtract $\delta$ from the transition time of all attributes whose influences did not change. The current time $t_c$ is updated from $t_i$ to $(t_i + \delta)$.
   (e) Simulate endogenous change from $(t_i + \delta)$ to $t_{i+1}$:
      i. Sample from each attribute's transition time, and choose the smallest, say $A_j$ with transition time $t$.
      ii. If $t + t_c > t_{i+1}$ then decrement transition times for all attributes by $t_{i+1} - t_c$, set $t_c$ to $t_{i+1}$, and return to step (2a) to process the next exogenous event.
      iii. Otherwise, attribute $A_j$ will transition in $t$ time units:
         - update the state to reflect $A_j$'s new value (the next larger or smaller)
         - re-compute influences based on $A_j$'s new value
         - decrement by $t$ the transition times for all attributes whose influences did not change. Advance the system time $t_c$ to $t_c + t$.
3. When the trial is generated,
   (a) Compute the weight for this trial on the basis of **W**.
   (b) Store the weight along with the values of all the elements collected because of **Q**.
   (c) Store the value, direction of change, and transition time estimates for all attributes at time $t_n + \delta$.

We have implemented this algorithm and applied it to the simple example presented in the paper:

1. Initially the patient has no internal bleeding or head injury; vital signs are normal and pupils are not dilated. (All this with probability 1.)
2. The first exogenous event is the collision at $t = 0$. We have a probability distribution over collision severity: mild, moderate, or severe.
3. The collision can cause internal bleeding and/or a head injury, with probabilities depending on its severity.
4. Head injury will endogenously cause pupils to dilate and vital signs to destabilize. Internal bleeding will also tend to destabilize vital signs (figures 2 and 3).
5. At $t = 10$ two observations are made: that vital signs are unstable and that pupils are not dilated. The first is always accurate, but the second can be incorrect: in a state where the pupils *are* dilated there is still a 10% chance that the report will incorrectly identify them otherwise.
6. We are interested in the likelihood of internal bleeding at $t = (10 + \delta)$ and also in the severity of the collision at $t = \delta$.

We show the results in Table 1: first the exact probability distribution over original collision severity and the exact distribution over internal bleeding values (both conditioned on the observations), then the imputed probabilities for increasing number of trials. For each run we report the number of trials, the time spent (in microseconds), the imputed distributions, and the "effective sample size" (ESS): a heuristic estimate of the importance or relevance of the samples that were gathered (i.e. a measure of the weight of the samples that were discarded for being incompatible with the observations, and equivalent to that number of samples drawn without weighting). These results should not be taken too seriously, since no attempt was made to optimize the code and the example is small. Still, we are encouraged that the technique can produce reasonable accuracy in times that are compatible with real-time decision making. Substantial work needs to be done to ascertain reasonable sample sizes for particular problems.

## 6 Related Work

Related work on probabilistic reasoning comes from the AI and statistics literature; our model of endogenous change draws on work from the AI literature on qualitative reasoning.

### 6.1 Temporal Reasoning with Probabilities

Several lines of research have attempted to extend classical AI approaches to temporal reasoning (which historically are defined in terms of a logical semantics) to a probabilistic framework. This work takes as fundamental a set of temporal or *diachronic* rules, describing relationships between the system's state at one time to its state at future times. *Synchronic* rules—those describing relationships between state variables at the same point in time—are generally absent from these models or handled as a special case. These models typically do not account for endogenous change in any deep way.

| Trials | Time (ms) | Collision Severity (mild, moderate, severe) | Bleeding (none, slight, gross) | ESS |
|---|---|---|---|---|
| Exact | | (0.240, 0.392, 0.368) | (0.070, 0.071, 0.859) | |
| 100 | 150 | (0.020, 0.580, 0.400) | (0.084, 0.126, 0.790) | 94 |
| 500 | 966 | (0.269, 0.465, 0.266) | (0.094, 0.071, 0.835) | 474 |
| 1000 | 1966 | (0.284, 0.405, 0.311) | (0.078, 0.061, 0.861) | 940 |
| 2500 | 4566 | (0.211, 0.437, 0.352) | (0.068, 0.088, 0.844) | 2357 |
| 5000 | 8833 | (0.232, 0.385, 0.383) | (0.074, 0.074, 0.852) | 4724 |
| 10000 | 18217 | (0.240, 0.387, 0.373) | (0.073, 0.071, 0.856) | 9459 |
| 50000 | 88967 | (0.248, 0.384, 0.368) | (0.069, 0.071, 0.860) | 47284 |
| 100000 | 179000 | (0.242, 0.397, 0.361) | (0.070, 0.071, 0.859) | 94542 |

Table 1: Empirical results for the three-event example

Dean and Kanazawa [1989] propose a "temporal belief network," a directed graphical model where nodes represent the truth of a state variable at a single point in time. The network is arranged into "time slices" representing the system's complete state at a single point in time, and time slices are duplicated over a predetermined and fixed-length time grid representing the time interval of interest. Links between state variables within a time slice (i.e. synchronic constraints) are disallowed. This network is formally equivalent to the network shown in Figure 1. There is no explicit model of endogenous change: judgments about the likelihood and nature of endogenous change are coded implicitly in the transition probabilities governing change from one time slice to the next.

Dean and Kanazawa suggest stochastic-simulation techniques for solving temporal belief networks, though subsequent work, e.g. by Kjærulff [1994], suggests methods for solving the network exactly. [Dean et al., 1992] suggests an extension of this framework to a class of semi-Markov models, but does not discuss an underlying model of endogenous change that would supply the needed transition-time probabilities.

Hanks and McDermott [1994] propose a similar model for temporal reasoning, this one explicitly based on a translation from symbolic diachronic rules to network structures. Once again synchronic constraints are disallowed and the model of endogenous change is captured in a set of transition probabilities directly assessing the likelihood of change over the intervals between exogenous events. The two main differences between this work and the Dean and Kanazawa work are first that Hanks and McDermott propose instantiating the network only at those points at which exogenous events or observations occur, whereas Dean and Kanazawa advocate instantiating the network on a fixed grid of time points, whether or not the system is likely to have changed in the interval. Second, Hanks and McDermott propose an algorithm for predicting future states of the system that involves instantiating the model *only* at those time points and for those attributes that are of interest to the decision maker: the model is built on demand, in response to user queries. On the other hand, the algorithm computes a *projected* probability only, so it lacks the ability to reason backward to compute the probability of an attribute at a prior time based on subsequent evidence.

### 6.2 Probabilistic Reasoning with Time

Another body of work, also commonly called probabilistic temporal reasoning, takes a fundamentally different approach to the problem. The general approach taken by [Provan, 1993], [Dagum and Galper, 1993], [Lekuona et al., 1995], [Berzuini et al., 1989], and others is to begin with a static probabilistic model of the system. In our example, the expert would be asked to assess a probability distribution over vital sign values conditioned on the existence of a head injury and/or internal bleeding, but without explicitly taking into account *when* either injury occurred.

Once the static probabilistic model is elicited, it is instantiated at various time points (e.g. at points at which observations are taken), and directed temporal links are drawn between propositions in one model to propositions in the temporally next model. There is no consensus at this point as to when the network should be instantiated, nor to exactly what links should be drawn. See [Provan, 1993] for a discussion.

This approach is strikingly different from ours: our model is built entirely from eliciting a diachronic or process-oriented model of the domain. Synchronic relationships are not directly elicited, but are inferred from the diachronic model. In the alternative view, synchronic relationships are directly elicited and the temporal relationships are inferred indirectly. The two approaches, and the resulting graphical structures, turn out to be extremely different in structure, and it is an unfortunate historical accident that the same set of terms ("dynamic belief networks," "temporal influence diagrams") are commonly used to describe both. See Figure 4 for a comparison of representative examples of the two sorts. The left is developed by Dean and Kanazawa, Hanks and McDermott, and our present work. The right is typical of networks built on the research summarized in this section.

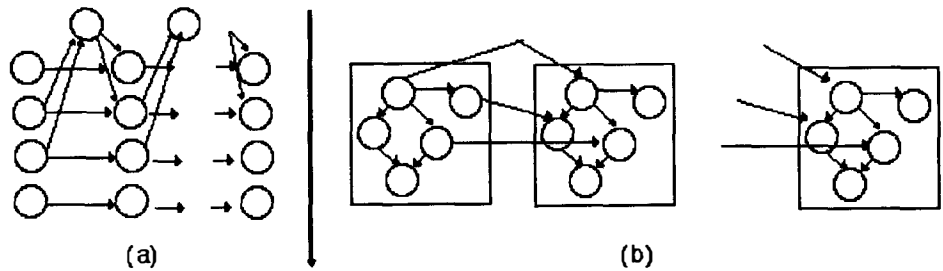

Figure 4: Two sorts of probabilistic temporal models

We suspect that the difference between the two approaches derives from assumptions made about how the models are to be elicited. In much of the latter work, e.g. [Dagum and Galper, 1993], it is assumed that significant parts of the model, most notably temporal relationships, are going to be inferred from data rather than elicited directly from an expert. The idea is that an algorithm will be presented with a sequence of "snapshots" of the system and will fit a descriptive model to that data. Thus static probabilistic information can be extracted from the snapshots, but the temporal structure of the system is not immediately observable to the learning system.

Our approach is to elicit the basic model structure from a human expert (though we hope to be able to capture or refine the model's probabilistic parameters empirically). That being the case the decision of what to model really amounts to what information the expert is most comfortable providing. We found that it was much easier for our domain expert to reason directly about the dynamics than to infer static relationships among state variables. Or more precisely, the expert generally made judgments about static relationships using an underlying dynamic model. Therefore it seemed more natural to elicit the dynamic model directly.

The existence of a dynamic or "process" model of the domain highlights the difference between diachronic and synchronic constraints: if one has a perfect dynamic model of the domain, any dependency between two state variables at a single point in time can be explained by a common event, and the dependency is captured there. Therefore no synchronic constraints are necessary. On the other hand, if *no* information about dynamics is available, dependency information *must* appear as static dependencies. Future work will address the problem of developing coherent and consistent hybrid models.

Complementary to this work, which is oriented toward *building* an appropriate network, is work oriented toward solving the network efficiently. See [Kjaerulff, 1994] for an exact solution and [Berzuini et al., 1994] for a Monte Carlo simulation technique that generalizes our sequential imputation approach.

### 6.3 Qualitative Reasoning

Our representation and reasoning techniques for endogenous change derive partly from AI work on Qualitative Reasoning [Dan Weld and de Kleer, 1989]: the abstracting of continuous attributes to qualitative values, the association of a direction of change with a variable, and the idea that some aspects of the system can cause others to change. There is a close similarity between the concept of a *process* in Qualitative Process Theory [Forbus, 1984] and our tables of exogenous influences: both dictate that sets of attributes being in a particular state will tend to exert upward or downward pressure on the values of other attributes. There are three main differences between the QP representation and our framework: a probabilistic model of exogenous change, an explicit model of what it means to observe the system and revise prior and subsequent beliefs as a result, and the association of numeric values (transition times) with endogenous influences.

Keeping numeric transition-time estimates is especially significant because it means we have a harder problem dealing with concordant and contrary influences: in most QP work the influences are either positive or negative, which means the change associated with a collection of positive influences is upwards (but not more strongly so) and if there are influences in both directions the net change is ambiguous.

Barahona [1994] notes this shortcoming of the QP approach and proposes a mixed qualitative/quantitative model for temporal reasoning. His model is similar to ours in that he can predict both the *time* to a state change as well as its direction; he also advocates a simple pooling mechanism for aggregating the effects of multiple processes (sum their individual influences). The differences include (1) his model allows reasoning with quantitative attributes instead of just qualitative abstractions, (2) his model has no provision for uncertainty about the state, the effects of exogenous events, or the nature or timing of endogenous change.

## 7 Conclusion

This work attempts to bridge a gap between probabilistic temporal reasoning work approached from the AI perspective and the work by the same name approached from the statistical perspective. The former tends to favor rule-based models for temporal reasoning, but the models of endogenous change have typically been weak. Key to the AI approach is that the model (rules and perhaps probabilistic parameters) are elicited from an expert. Typically diachronic relationships (those relating the system's state at one time to its state at some subsequent time) have been more common than explicit synchronic relationships (relating one aspect of the system's state to another at a single point in time). In general we found our expert more comfortable talking about the evolution of the patient's state over time than about single-state relationships. (More accurately, he would be willing to think about synchronic relationships, but most often would use diachronic relationships to do so.)

The common modeling method in statistical approaches to the problem would be to elicit a synchronic (static) model of the domain, then fit temporal relationships among these static models.

Although the preliminary results look encouraging, we need to work on several aspects of the model. In particular the method for reasoning about and combining multiple influences is fairly arbitrary, and should at least be tested in a variety of different problem domains. Eventually we would like to come up with a variety of techniques for eliciting endogenous models that could be used as appropriate for a given domain or physical reality, then plug the resulting model into our more generic simulation framework.